\documentclass{article} 
\usepackage{iclr2027_conference,times}

\usepackage{amsmath,amsfonts,bm}

\def\eqref#1{equation~\ref{#1}}

\def\1{\bm{1}}

\DeclareMathAlphabet{\mathsfit}{\encodingdefault}{\sfdefault}{m}{sl}
\SetMathAlphabet{\mathsfit}{bold}{\encodingdefault}{\sfdefault}{bx}{n}

\usepackage{hyperref}
\usepackage{url}
\usepackage{amssymb}
\usepackage[dvipsnames]{xcolor}
\usepackage{booktabs}
\usepackage{enumitem}
\usepackage{wrapfig}
\usepackage{colortbl}
\usepackage{multirow}
\usepackage{tabularx}
\usepackage{pifont}
\usepackage[most]{tcolorbox}
\usepackage{caption}
\usepackage{wrapfig}
\usepackage{enumitem}

\newtcolorbox{bluebox}[1][]{
  enhanced,
  colframe=blue!40!gray,
  colback=white,
  coltitle=white,
  colbacktitle=blue!40!gray,
  width=\linewidth,
  arc=2mm,
  auto outer arc,
  boxrule=0.5pt,
  left=10pt,
  right=10pt,
  drop shadow={black!50!white},
  top=10pt,
  bottom=10pt,
  title={#1},
  fonttitle=\bfseries,
  title code={\node[rounded corners, fill=blue!75!black, draw=none, text=white] at (frame.title) {\textbf{#1}};},
  attach boxed title to top center={yshift=-2mm},
  boxed title style={sharp corners, size=small}
}

\title{Save Your Saturated Data: Learning Beyond Reward Saturation in Group-Based RL}

\author{Ziyuan Yang\thanks{equal contribution} \ \ \ Yike Wang\footnotemark[1] \ \ \ Shangbin Feng\footnotemark[1] \ \ \ Yulia Tsvetkov \\
University of Washington \\
\texttt{ziyuan86@uw.edu} \ \ \ \texttt{\{yikewang, shangbin\}@cs.washington.edu}} 

\iclrfinalcopy 
\begin{document}

\maketitle

\begin{abstract}

Group-relative reinforcement learning (RL) relies on reward variation among sampled responses to estimate informative relative advantages. As language models become increasingly capable, existing training data can become \emph{reward-saturated}: all sampled responses to the same problem might receive equally high rewards, where the group-relative learning signals vanish and leave previously useful data obsolete. In this work, we investigate \emph{whether useful learning signals can be recovered from such saturated data}. We study interventions at four levels of group-policy RL pipelines---data, rollout, reward, and advantage---and conduct extensive RL training on saturated reasoning data \emph{only}. While standard GRPO on saturated data would almost always yield near-0 advantages and near-noise signals, diverse interventions successfully recycle and repurpose such data: among the proposed strategies, interventions at rollout generation are consistently most effective: nudging the policy to generate ``high-quality'', incorrect solutions introduces rollouts with poor rewards into saturated groups as negative samples, which turns out to improve GRPO by 6.4\% to 9.0\% across Qwen3-1.7B and 4B. 
Other interventions such as increasing rollout temperature or adding auxiliary rewards can also restore non-zero advantages, but yield less consistent gains. Further analyses show that effective negative rollouts require informative negative trajectories, that the method remains effective alongside unsaturated data, and that it supports iterative recycling of newly saturated examples. While increasingly stronger LLMs would render more data as saturated, our results demonstrate that \emph{don't waste your saturated data}: with the right strategies they can be recycled into useful RL training signals in an increasingly data-scarce world. Our code is available at \href{https://github.com/Ziyuan-Yang/saturatedRL}{https://github.com/Ziyuan-Yang/saturatedRL}.


\end{abstract}

\section{Introduction}

Group-based reinforcement learning (RL) has become a key component in LLM post-training, especially when used with verifiable rewards. Methods such as GRPO~\citep{shao2024deepseekmathpushinglimitsmathematical, deepseekr1} and DAPO~\citep{yu2025dapo} sample multiple responses for a prompt, calculate rewards, and estimate advantage based on group reward differences. As a result, obtaining useful learning signals is contingent on having responses of diverse quality and reward values.

However, this learning signal vanishes at both extremes: When a problem is hard and all sampled rollouts are incorrect, uniformly low rewards won't leave meaningful advantage signals to train on. Prior work has primarily explored this problem through adaptive exploration~\citep{zhang2026aero, huang26multiVIGOR, agrawal2026offcontext}, advantage modification~\citep{le2026no, he2026advantage}, and self-evolving curricula~\citep{huang2026rzero, zhao2025absolute}. On the contrary, when a problem is easy and all rollouts are correct, uniformly high rewards suffer from the same advantage vanishing problem. We focus on the latter scenario, which we refer to as \emph{reward saturation}, especially timely as LLM progress far exceeds the availability of challenging training data.

This reflects a growing tension between model capability and data utility: as the policy improves, more of its existing training data may become reward-saturated. While we can always curate harder problem sets and replace existing data, it is not sufficient alone: constructing high-quality challenging problems often requires substantial expert effort, reliable verification, and careful control over the problem distribution~\citep{parashar2026curriculum, huang2026rzero, bao2026questionbeget}. Our work asks a complementary question: \emph{Can we recycle training data with saturated rewards for useful learning signals in group-based RL?}

While previous work \citep{liang2026correctlearnreinforcementlearning} offered a preliminary step of intervening with rollout generation, we systematically investigate this research question through interventions at four levels of GRPO: \emph{data} (e.g., adding irrelevant context to the training problem), \emph{rollout} (e.g., constructing wrong negative rollouts), \emph{reward} (e.g., adding additional fine-grained rewards), and \emph{advantage} (e.g., modifying the advantage estimation function). We conduct extensive RL training across models of varying sizes and families, using these proposed interventions on reward-saturated data, and evaluate the trained policies across eight datasets spanning math, reasoning, and instruction following.

Across these interventions, we find that reward-saturated data can indeed be recycled into useful training signals. Among them, negative rollout is consistently the most effective: using the policy itself to generate incorrect solutions introduces informative trajectory-level contrasts within group, improving standard GRPO by 9.0\% on Qwen3-1.7B and 6.4\% on Qwen3-4B. Other interventions, like higher-temperature sampling, reward shaping, and advantage manipulation, can also recover learning signals, but yield smaller or less consistent gains. Further analyses show that effective negative rollouts require informative unsuccessful trajectories, remain effective when saturated and unsaturated data are mixed, enable new saturated examples to be iteratively recycled as the policy improves, and modifying the advantage function  
alone does not lead to better performance. 

Taken together, our results suggest that reward saturation does not have to mark the end of a training example's utility in group-based RL. As models become stronger and increasingly more data become saturated, these saturated data can be continually recycled into useful RL training signals rather than discarded, enabling continued policy improvement even as saturation grows.

\begin{figure}[t]
\centering
\includegraphics[width=0.9\linewidth]{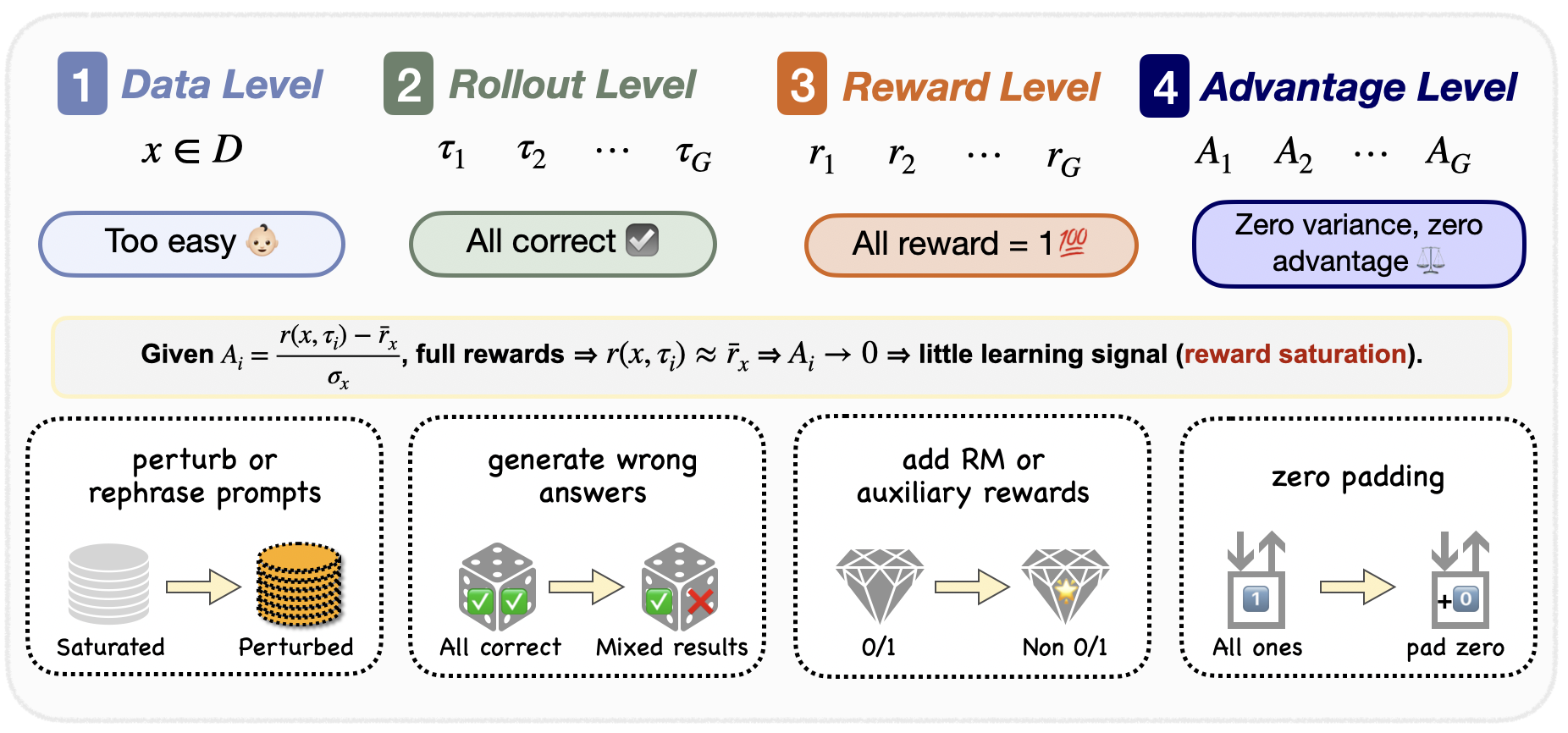}
\caption{Overview of our work. (Top) Reward saturation in GRPO: When training data are easy, all \textit{G} rollouts could be correct and receive the maximum reward, leading to zero advantage and thus little policy update. (Bottom) Multi-Level Interventions: We intervene at one of four stages of GRPO training: data, rollout, reward, and advantage to restore informative learning signals from reward-saturated examples. These interventions act at different stages of the training pipeline but share the same goal: restoring a usable group-relative learning signal for otherwise saturated prompts.}
\label{fig:main_overview}
\end{figure}

\section{Related Work}

\paragraph{Group-Based Reinforcement Learning with Verifiable Rewards.} 
Reinforcement learning with verifiable rewards (RLVR) has emerged as an effective paradigm for improving LLM reasoning. Group-based methods such as GRPO~\citep{shao2024deepseekmathpushinglimitsmathematical, deepseekr1} and DAPO~\citep{yu2025dapo} estimate relative advantages from multiple rollouts of the same prompt, relying on within-group reward variation for learning. When all responses receive identical rewards, group-relative advantages collapse, motivating our work on intervention on four levels of policy training.

\paragraph{Recovering Learning Signals from Low-Information Groups.} 
Prior work largely studies the difficult data regime, where most or all rollouts are incorrect. Approaches include filtering overly hard questions~\citep{yu2025dapo}, improving exploration~\citep{zhang2026aero,huang26multiVIGOR,agrawal2026offcontext}, modifying advantage estimation~\citep{le2026no,he2026advantage}, adapting the training distribution~\citep{huang2026rzero,zhao2025absolute}, and constructing harder problems~\citep{parashar2026curriculum,huang2026rzero,bao2026questionbeget}. In contrast, we study the opposite: prompts with successful rollouts, where group-relative advantages likewise collapse.


\paragraph{Learning from Reward-Saturated Data.}
Recent work has begun to directly address reward saturation. Diagnostic studies show that advantage collapse strongly predicts training stagnation~\citep{he2026advantage}. Existing mitigations either avoid uninformative groups through pre-rollout difficulty estimation~\citep{hu2026vade}, reuse previously effective groups through off-policy replay~\citep{mao2026popo}, or recover within-group variation through virtual reward samples~\citep{he2026advantage}, auxiliary trajectory-quality signals~\citep{deng2026prismgrpo}, or constrained exploratory sampling~\citep{liang2026correctlearnreinforcementlearning}. We extend this direction by systematically studying interventions across four stages of RL---\emph{data}, \emph{rollout}, \emph{reward}, and \emph{advantage}---to examine which recovered signals support effective learning, with our negative-rollout intervention making this distinction explicit.

\section{Method}

\subsection{GRPO Preliminary}

GRPO~\citep{shao2024deepseekmathpushinglimitsmathematical} is a policy optimization algorithm that has demonstrated strong performance on various reasoning tasks. Let $x$ denote a prompt sampled from dataset $\mathcal{D}$, and let $\{\tau_i\}_{i=1}^{G}$ denote a group of $G$ rollouts generated by the current policy. Each rollout $\tau_i$ receives a binary verifiable reward $r(x,\tau_i)\in\{0,1\}$, determined by whether its final answer is verified as correct against the ground-truth answer. GRPO then computes the group-level reward mean $\bar{r}_x$ and standard deviation $\sigma_x$ to obtain the normalized group-relative advantage:

\vspace{-5pt}
\begin{equation}
A(x,\tau_i)
=
\frac{r(x,\tau_i)-\bar{r}_x}{\sigma_x}.
\end{equation}

For simplicity, we denote $A(x,\tau_i)$ as $A_i$. Let $\pi_\theta$ denote the optimized policy and $\pi_{\mathrm{ref}}$ denote a frozen reference policy. GRPO adopts a PPO-style clipped objective with token-level importance ratio $\rho_{i,t}(\theta)$ and KL regularization weight $\beta$:

\vspace{-10pt}
\begin{equation}
\small
\begin{aligned}
\mathcal{J}(\theta)
=
\mathbb{E}_{x\sim\mathcal{D},\tau_i\sim\pi_{\theta_{\mathrm{old}}}}
\left[
\sum_{t=1}^{|\tau_i|}
\min
\left(
\rho_{i,t}(\theta)A_i, \operatorname{clip}
(\rho_{i,t}(\theta),1-\epsilon,1+\epsilon)A_i
\right)
\right]
-\beta
\mathbb{E}_{x\sim\mathcal{D}}
\left[
D_{\mathrm{KL}}
(\pi_\theta\|\pi_{\mathrm{ref}})
\right].
\end{aligned}
\label{eq:grpo}
\end{equation}
\vspace{-10pt}

\subsection{Intervention at Different Levels}

We define a rollout group as \emph{reward-saturated} when all responses in the group receive the same maximum reward, such that the group-relative advantages vanish. In our setting, this corresponds to all $G$ sampled responses being correct. As discussed in Eq.~\ref{eq:grpo}, GRPO relies on reward differences among multiple rollouts sampled from the same prompt to construct group-relative advantages. When all sampled trajectories receive identical rewards, the within-group reward variance collapses, resulting in uninformative relative advantages and limited policy updates. 

To restore useful learning signals for such saturated groups, we investigate interventions at four stages of the GRPO pipeline: \emph{data}, \emph{rollout}, \emph{reward}, and \emph{advantage}. For each intervention, we first identify reward-saturated prompts by sampling $G$ rollouts and retaining prompts for which all $G$ responses are correct, and then apply the intervention to their saturated rollout groups. Although operating at different stages of training, all interventions share the same goal of restoring usable group-relative learning signals from saturated prompts. We next describe each intervention in turn.

\paragraph{Data-Level Intervention}

At the data level, we perturb the original training data to induce greater variation in rollout outcomes and thereby reduce reward saturation. We consider two variants:
\begin{itemize}
    \item \textit{Irrelevant Information:} augmenting each prompt with irrelevant information, yielding $D_{\mathrm{irrelevant}} = \mathrm{Augment}_{\mathrm{irr}}(D)$.
    \item \textit{Prompt Rephrasing:} rephrasing each prompt while preserving its underlying semantics, yielding $D_{\mathrm{rewrite}} = \mathrm{Rewrite}(D)$.
\end{itemize}

We investigate whether these data-level perturbations can reduce saturation by inducing greater variation in rollout outcomes and, consequently, within-group rewards.


\paragraph{Rollout-Level Intervention}

At the rollout level, we modify the generation process to increase reward heterogeneity within each rollout group. We consider two general strategies:
\begin{itemize}
    \item \textit{Higher Temperature:} increasing the sampling temperature $t$ to encourage more diverse trajectories and increase the likelihood of obtaining both correct and incorrect rollouts.
    \item \textit{Negative Rollout:} explicitly constructing incorrect rollouts for saturated groups in which all sampled responses receive the maximum verification reward.
\end{itemize}

For negative rollout construction, let $\mathcal{G}_{\mathrm{correct}}$ and $\mathcal{G}_{\mathrm{wrong}}$ denote the sets of correct and constructed incorrect rollouts, respectively. We form the rollout group as $\mathcal{G} = \mathcal{G}_{\mathrm{correct}} \cup \mathcal{G}_{\mathrm{wrong}}$.

To construct $\mathcal{G}_{\mathrm{wrong}}$, we append an instruction to the original prompt that explicitly asks the current policy to generate an incorrect solution; the detailed prompt is provided in Appendix~\ref{appendix:wrong_answer_suffix}. We further investigate alternative negative rollout construction strategies in Section~\ref{analysis:wrong_answer}.

\paragraph{Reward-Level Intervention}

At the reward level, we investigate whether alternative reward signals can alleviate saturation caused by coarse-grained verification rewards. Binary verification provides no distinction among trajectories once they all satisfy the correctness criterion. We therefore augment the verification reward with auxiliary signals that may differentiate otherwise equally rewarded trajectories. We consider three auxiliary reward signals, each evaluated independently:

\begin{itemize}
    \item \textit{Reward Model:} a reward-model score $R_{\mathrm{RM}}(\tau_i)$ assigned to each rollout $\tau_i$ by a pretrained reward model.
    \item \textit{Reasoning Quality:} an LLM-as-a-judge score $R_{\mathrm{Reason}}(\tau_i)$ that evaluates the correctness of the reasoning process and intermediate steps in each rollout.
    \item \textit{Response Diversity:} an LLM-as-a-judge score $R_{\mathrm{Diversity}}(\tau_i)$ that measures the distinctiveness of each rollout's reasoning approach relative to the other $G-1$ rollouts in the same group, assigning higher scores to more distinct trajectories.
\end{itemize}

For each intervention, we define the augmented reward as
\begin{equation}
R(\tau_i)
= R_{\mathrm{ver}}(\tau_i) + \lambda R_k(\tau_i),
\qquad
R_k \in
{
R_{\mathrm{RM}},
R_{\mathrm{Reason}},
R_{\mathrm{Diversity}}
},
\end{equation}

where $R_{\mathrm{ver}}$ denotes the original binary verification reward and $\lambda$ controls the contribution of the auxiliary reward. Each $R_k$ is evaluated separately rather than combining multiple auxiliary signals. These interventions test whether increasing reward granularity can restore informative group-relative advantages under reward saturation.


\paragraph{Advantage-Level Intervention}

One might argue that, since all of these rollouts achieve perfect reward, they should be treated equally with positive rewards.
We investigate whether modifying advantage estimation can achieve this at the advantage level. We consider \textit{Zero-Padding}, which appends $k$ zero-reward entries to the original reward group when computing group-relative advantages. These additional entries increase the within-group reward variance, yielding non-zero advantages for the original successful rollouts.


Formally, given an original reward group $\{r(x,\tau_i)\}_{i=1}^{G}$, Zero-Padding augments it with $k$ zeros and recomputes the group statistics:

\begin{equation}
\begin{aligned}
A'(x,\tau_i)
&=
\frac{r(x,\tau_i)-\bar{r}'_x}{\sigma'_x}, &
\bar{r}'_x = \mathrm{mean}
\left(
\{r(x,\tau_i)\}_{i=1}^{G}
\cup
\{\underbrace{0,\dots,0}_{k}\}
\right),
\end{aligned}
\label{eq:advantage_zero_padding}
\end{equation}

where $\sigma'_x$ denotes the standard deviation of the augmented reward group.

Zero-Padding restores non-zero relative advantages without modifying the training data, rollout trajectories, or reward function, while preserving the same positive preference across all rollouts. It therefore serves as a controlled intervention that increases the magnitude of the optimization signal without introducing new information.

\section{Experiment}

\subsection{Experiment Settings}
We conduct RL training on a $4.3\mathrm{K}$ subset of the MATH dataset~\citep{hendrycks2021measuring}. To construct a reward-saturated setting, we retain only problems for which all eight independently sampled rollouts from the base model are correct. To evaluate both in-domain mathematical reasoning and broader generalization, we consider eight benchmarks spanning mathematical reasoning, general reasoning, and instruction following: MATH-500, Minerva~\citep{lewkowycz2022solving}, BBH~\citep{suzgun-etal-2023-challenging}, GPQA-Diamond (GPQA)~\citep{rein2024gpqa}, AIME24~\citep{aime_1983_2024}, AIME25, IFBench~\citep{pyatkin2025generalizing}, and IFEval~\citep{zhou2023instructionfollowingevaluationlargelanguage}.

We use the Qwen3 series~\citep{yang2025qwen3technicalreport} as the primary backbone, including the 1.7B and 4B variants with the thinking mode disabled. All experiments are conducted on NVIDIA H200 GPUs using the \textit{verl} framework~\citep{sheng2024hybridflow}. Models are trained for two epochs with a rollout group size of $G=8$. For negative rollout construction, we set $G_{\mathrm{wrong}}=2$ while keeping the total group size fixed. Additional implementation details are provided in Appendix~\ref{appendix:Experimental_details}. 

\begin{table*}[t]
\centering
\caption{Performance comparison across eight benchmarks using Qwen3-1.7B and Qwen3-4B. We compare the base model, standard GRPO, Mixed-CUTS, SFT,  SFT $\rightarrow$ GRPO and interventions at the data, rollout, reward, and advantage levels. We report avg@32 for AIME24 and AIME25 and pass@1 for all other benchmarks. The best and second-best results within each model group are highlighted in \textbf{bold} and \underline{underline}, respectively.}
\label{tab:math_benchmark}
\resizebox{\textwidth}{!}{
\begin{tabular}{lccccccccc}
\toprule[1.75pt]
\textbf{Settings} & \textbf{MATH500} & \textbf{GPQA} & \textbf{BBH} &  \textbf{Minerva} & \textbf{AIME24} & \textbf{AIME25} & \textbf{IFBench} & \textbf{IFEval} & \textbf{Average} \\
\midrule[1pt]
\multicolumn{9}{l}{\textit{\textbf{Policy: Qwen3-1.7B}}} \\
\addlinespace[2pt]
Base Model & 60.00 & 25.80 & 9.65 & 18.00 & 11.46 & 9.06 & 23.20 & 68.00 & 28.15 \\
\midrule
\multicolumn{10}{l}{\textit{Baselines}} \\
GRPO & 62.60 & 24.24 & 10.86 & \underline{17.28} & 17.08 & 9.17 & 21.20 & 69.00 & 28.93 \\
Mixed-CUTS & 61.00 & \textbf{32.32} & 10.55 & 18.38 & 16.77 & \underline{12.81} & \underline{25.20} & \textbf{70.20} & \underline{30.91} \\
SFT & 58.40 & 29.29 & 8.23 & 18.01 & 6.67 & 10.10 & 22.00 & 67.40 & 27.51 \\
SFT$\rightarrow$GRPO & 61.20 & 20.20 & 8.89 & 16.91 & 15.42 & 10.52 & 22.40 & 65.40 & 27.62 \\
\midrule
\multicolumn{10}{l}{\textit{Data Level}} \\
Irrelevant Information & 62.20 & 20.71 & 10.04 & 16.18 & 16.56 & 7.50 & 22.40 & 68.80 & 28.05 \\
Prompt Rephrasing & 63.00 & 19.19 & 9.67 & 15.44 & 12.19 & 10.63 & 23.20 & 66.00 & 27.41 \\
\midrule
\multicolumn{10}{l}{\textit{Rollout Level}} \\
Higher temperature & 63.60 & 25.25 & 10.40 & 17.28 & 16.25 & 12.71 & 24.00 & 68.40 & 29.74 \\
Negative Rollout & \underline{64.80} & \underline{29.80} & \textbf{13.01} & \textbf{18.38} & \textbf{20.31} & \textbf{14.38} & 24.00 & 67.60 & \textbf{31.53} \\
\midrule
\multicolumn{10}{l}{\textit{Reward Level}} \\
Reward Model & 62.00 & 28.28 & 10.29 & \textbf{18.38} & \underline{18.85} & 9.38 & 22.00 & 67.80 & 29.62 \\
Reasoning Quality & \textbf{65.80} & 27.27 & 10.40 & 16.54 & 16.88 & 10.42 & 23.60 & 69.20 & 30.01 \\
Response Diversity & 61.20 & 24.24 & 10.33 & 16.54 & 10.31 & 10.52 & 22.80 & 67.80 & 27.97 \\
\midrule
\multicolumn{10}{l}{\textit{Advantage Level}} \\
Zero Padding ($k=2$) & 62.80 & 27.27 & 10.84 & 16.54 & 16.04 & 10.52 & 22.40 & 66.80 & 29.15 \\
Zero Padding ($k=3$) & 64.00 & 28.28 & \underline{10.93} & 16.54 & 18.33 & 12.50 & \textbf{25.60} & \underline{69.40} & 30.70 \\
\midrule[1pt]
\addlinespace[2pt]
\multicolumn{9}{l}{\textit{\textbf{Policy: Qwen3-4B}}} \\
\addlinespace[2pt]
Base Model & 67.20 & 38.89 & 15.00 & 23.90 & 19.58 & 13.13 & 28.40 & 80.00 & 35.76 \\
\midrule
\multicolumn{10}{l}{\textit{Baselines}} \\
GRPO & \underline{70.60} & 31.31 & 16.13 & \underline{25.74} & 31.98 & \textbf{27.40} & 28.00 & 78.80 & 38.74 \\
Mixed-CUTS & 70.40 & 38.89 & 16.28 & 24.26 & 23.33 & 10.73 & 29.60 & 80.80 & 36.79 \\
SFT & 67.20 & 38.89 & 13.69 & 23.90 & 16.04 & 14.06 & 30.00 & 81.20 & 35.62 \\
SFT$\rightarrow$GRPO & 69.40 & 31.31 & 14.93 & 25.37 & 30.21 & 19.69 & 27.20 & 82.80 & 37.61 \\
\midrule
\multicolumn{10}{l}{\textit{Data Level}} \\
Irrelevant Information & \textbf{71.00} & \underline{40.40} & 15.80 & 25.37 & 27.92 & 25.21 & 28.00 & 80.40 & 39.26 \\
Prompt Rephrasing & 68.60 & 34.34 & 15.93 & 24.63 & 22.08 & 23.23 & 29.60 & 80.40 & 37.35 \\
\midrule
\multicolumn{10}{l}{\textit{Rollout Level}} \\
Higher temperature & 69.40 & 38.38 & 16.31 & \textbf{26.10} & 23.85 & 18.13 & 29.60 & \textbf{82.60} & 38.05 \\
Negative Rollout & \underline{70.60} & \underline{40.40} & \textbf{18.27} & \textbf{26.10} & \underline{33.02} & 25.52 & \textbf{33.20} & \textbf{82.60} & \textbf{41.22} \\
\midrule
\multicolumn{10}{l}{\textit{Reward Level}} \\
Reward Model & 70.20 & \textbf{41.92} & \underline{17.83} & 22.79 & \textbf{36.46} & 20.73 & 31.20 & 80.80 & \underline{40.24} \\
Reasoning Quality & \textbf{71.00} & 32.32 & 15.75 & \underline{25.74} & 23.02 & \underline{27.19} & 30.80 & 80.00 & 38.23 \\
Response Diversity & 68.20 & 34.34 & 15.29 & 23.53 & 13.44 & 12.92 & \underline{31.20} & \underline{82.00} & 35.11 \\
\midrule
\multicolumn{10}{l}{\textit{Advantage Level}} \\
Zero Padding ($k=2$) & 62.80 & 31.82 & 13.83 & 19.85 & 14.06 & 11.88 & 27.60 & 81.40 & 32.90 \\
Zero Padding ($k=3$) & 67.20 & 30.30 & 13.50 & 22.79 & 20.21 & 23.13 & 30.00 & 79.80 & 35.87 \\

\bottomrule[1.75pt]
\end{tabular}}
\end{table*}

\subsection{Baselines}


We compare our approach with representative baselines covering pretrained models, standard RL training, and prior methods for saturated RL data: 

\begin{itemize}[topsep=1pt, itemsep=2pt, parsep=0pt, partopsep=0pt]
    \item \textbf{Base Model.} The pretrained backbone model without post-training.
    \item \textbf{GRPO.} The standard GRPO algorithm, serving as the primary RL baseline.
    \item \textbf{Mixed CUTS}~\citep{liang2026correctlearnreinforcementlearning}. A baseline that constructs exploratory rollouts after warmup tokens by uniformly sampling to reduce saturation.
    \item \textbf{SFT.} Supervised fine-tuning on solutions generated by a teacher model (Qwen3-8B).
    \item \textbf{SFT $\rightarrow$ GRPO.} Supervised fine-tuning followed by standard GRPO training.
\end{itemize}
\vspace{-5pt}


\section{Results}

Table~\ref{tab:math_benchmark} reports results on saturated training data across two Qwen3 backbones.

\paragraph{Standard training baselines provide limited gains on saturated data.}
Standard GRPO yields only modest improvements over the base model, increasing the average score from 28.15 to 28.93 on Qwen3-1.7B and from 35.76 to 38.74 on Qwen3-4B. SFT alone provides little benefit over the base model, with average scores slightly decreasing from 28.15 to 27.51 on Qwen3-1.7B and from 35.76 to 35.62 on Qwen3-4B. Further applying GRPO after SFT reaches 27.62 and 37.61, respectively, but still falls short of standard GRPO (28.93 and 38.74). Mixed-CUTS is competitive on the smaller model, reaching 30.91, but drops below GRPO on Qwen3-4B (36.79). Overall, standard imitation-based training and existing strategies provide limited or inconsistent improvements when training data are already reward-saturated.

\paragraph{Negative rollouts are the most effective intervention under reward saturation.}
Among all evaluated strategies, negative rollout construction achieves the strongest overall performance on both model sizes. For Qwen3-1.7B, negative rollouts improve the average score from 28.93 with standard GRPO to 31.53, yielding a +2.60 absolute gain (+9.0\% relative improvement) and outperforming Mixed-CUTS (30.91). For Qwen3-4B, the average score improves from 38.74 to 41.22, corresponding to a +2.48 absolute gain (+6.4\% relative improvement). The gains are also pronounced on challenging mathematical reasoning benchmarks; for example, AIME25 improves from 9.17 to 14.38 on Qwen3-1.7B. These results show that reward-saturated examples need not be discarded: constructing informative negative trajectories introduces meaningful within-group reward contrast and restores useful learning signals for GRPO.

\paragraph{Increasing diversity alone provides limited and inconsistent improvements.}
Interventions that perturb the input or increase rollout diversity provide smaller and less consistent gains. Increasing the rollout temperature improves the average score from 28.93 to 29.74 on Qwen3-1.7B, but decreases it from 38.74 to 38.05 on Qwen3-4B. Data-level perturbations show a similar pattern: irrelevant information improves Qwen3-4B to 39.26 but provides no gain on Qwen3-1.7B, while prompt rephrasing underperforms GRPO at both scales. These results suggest that increasing diversity alone does not consistently overcome reward saturation; what matters is whether the generated trajectories introduce informative reward differences within each rollout group.

\paragraph{Reward shaping and advantage manipulation recover weaker signals.}
Reward-level and advantage-level interventions can partially restore learning signals, but remain less consistent than negative rollouts. Reward-model supervision is particularly effective on Qwen3-4B, improving the average score from 38.74 to 40.24, although it remains below negative rollouts at 41.22. In contrast, reasoning-quality and response-diversity evaluators do not consistently improve over standard GRPO. Zero-reward padding improves Qwen3-1.7B, reaching 30.70 with three additional zero-reward samples, but substantially underperforms GRPO on Qwen3-4B. Together, these results suggest that simply introducing non-zero advantages or finer-grained reward signals can help, but is less reliable than constructing meaningful trajectory-level contrasts.

\section{Analysis}

\subsection{Effective Negative Rollouts Require Informative Trajectories}
\label{analysis:wrong_answer}

We investigate whether the benefit of negative rollouts arises simply from restoring reward variance, or whether the construction of the negative trajectories also matters. We compare three strategies that differ in how coherently the negative outcome is reflected throughout the trajectory: (1) \textit{answer replacement}, which replaces only the final answer of an originally correct rollout while preserving its reasoning trajectory; (2) \textit{wrong-answer-conditioned continuation}, which provides a reasoning prefix and conditions the model to continue toward a predetermined incorrect answer; and (3) \textit{model-generated negative rollout}, our default strategy, which instructs the current policy to generate an incorrect solution from scratch. All three strategies restore reward variation within saturated groups, but differ in how faithfully the resulting trajectories represent unsuccessful generation.

As shown in Table~\ref{tab:wrong_answer}, simply restoring reward variation does not consistently improve performance. Answer replacement performs worse than standard GRPO, with average scores of 27.70 on Qwen3-1.7B and 33.38 on Qwen3-4B. Wrong-answer-conditioned continuation improves over answer replacement, reaching 30.29 and 37.31, respectively, but remains below GRPO on Qwen3-4B. In contrast, model-generated negative rollouts achieve the strongest average performance on both backbones, reaching 31.53 on Qwen3-1.7B and 41.22 on Qwen3-4B.

These results show that introducing negative rollouts alone is not sufficient: \emph{how the corresponding negative trajectories are constructed matters}. Answer replacement creates a mismatch between an otherwise successful reasoning trajectory and an artificially incorrect final answer, while conditioned continuation constrains only part of the generation process. In contrast, model-generated negative rollouts allow the policy to construct an entire unsuccessful trajectory, producing a more coherent correspondence between the generated trajectory and its negative reward.

\begin{table*}[h]
\centering
\caption{Comparison of negative rollout construction strategies across eight benchmarks. Although all three strategies restore reward variation within saturated groups, model-generated negative rollouts achieve the strongest average performance across both model scales.}
\vspace{-5pt}
\vspace{1mm}
\label{tab:wrong_answer}
\resizebox{\textwidth}{!}{
\begin{tabular}{lccccccccc}
\toprule[1.75pt]
\textbf{Method} & \textbf{MATH500} & \textbf{GPQA} & \textbf{BBH} & \textbf{Minerva}  &  \textbf{AIME24} & \textbf{AIME25} & \textbf{IFBench} & \textbf{IFEval} & \textbf{Average} \\
\midrule[1pt]
\multicolumn{9}{l}{\textit{\textbf{Qwen3-1.7B}}} \\
\midrule
GRPO & 62.60 & 24.24 & 10.86 & 17.28 & 17.08 & 9.17 & 21.20 & 69.00 & 28.93 \\
\midrule
Answer Replacement & 59.20 & 27.78 & 13.10 & 14.71 & 2.50 & 7.29 & 29.20 & 67.80 & 27.70 \\
Wrong-Answer-Cond. & 64.20 & 29.80 & 10.86 & 17.28 & 17.19 & 9.17 & 23.20 & 70.60 & 30.29 \\
Model-Generated Negative & 64.80 & 29.80 & 13.01 & 18.38 & 20.31 & 14.38 & 24.00 & 67.60 & \textbf{31.53} \\
\midrule
\multicolumn{9}{l}{\textit{\textbf{Qwen3-4B}}} \\
\midrule
GRPO & 70.60 & 31.31 & 16.13 & 25.74 & 31.98 & 27.40 & 28.00 & 78.80 & 38.74 \\
\midrule
Answer Replacement & 54.20 & 39.90 & 16.37 & 23.16 & 15.42 & 3.96 & 34.80 & 79.20 & 33.38 \\
Wrong-Answer-Cond. & 71.20 & 36.36 & 16.86 & 25.37 & 24.48 & 13.65 & 32.00 & 78.60 & 37.31 \\
Model-Generated Negative & 70.60 & 40.40 & 18.27 & 26.10 & 33.02 & 25.52 & 33.20 & 82.60 & \textbf{41.22} \\
\bottomrule[1.75pt]
\vspace{-5pt}
\end{tabular}}
\end{table*}

\subsection{Negative Rollouts Generalize Beyond Fully Saturated Data}
\label{analysis:mixed_data}


Our main experiments focus on fully saturated training data, while saturated and non-saturated groups may coexist in practice. We therefore evaluate Negative Rollout under different saturation ratios. Before training, we identify saturated prompts and partition the data into saturated and non-saturated pools, from which we construct training sets with varying proportions. Negative Rollout is applied only to saturated groups, while non-saturated groups follow standard GRPO. This setting allows us to evaluate whether the method remains effective when reward saturation affects only part of the training data.

\begin{figure}[t] 
    \centering
    \vspace{-10pt}
    \includegraphics[width=1\linewidth]{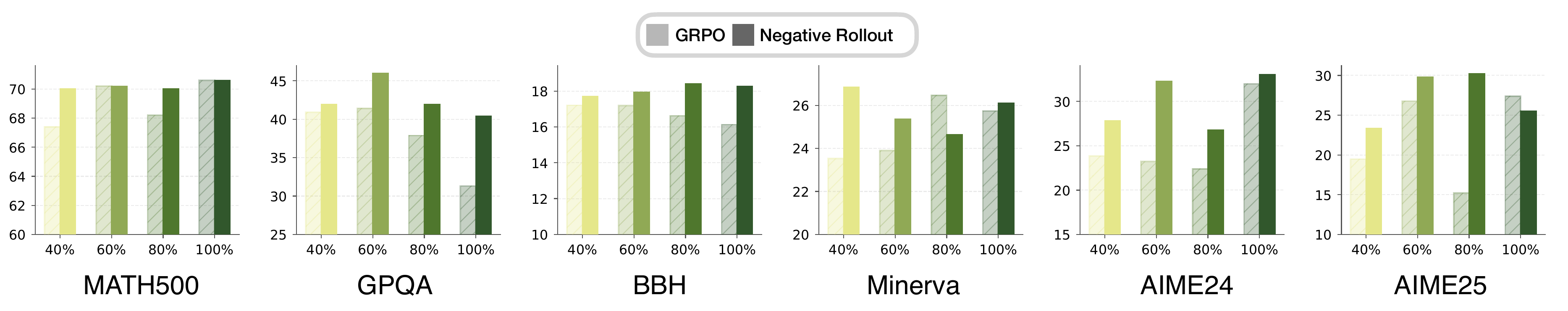}
    \caption{Performance of GRPO and Negative Rollout across different ratios of reward-saturated training data. Negative Rollout improves aggregate performance across all saturation ratios, with gains varying across benchmarks.}
    \vspace{-10pt}
    \label{fig:saturatio_data_ratio}
    \vspace{-5pt}
\end{figure}

As shown in Figure~\ref{fig:saturatio_data_ratio}, Negative Rollout improves the average performance over GRPO across all tested saturation ratios, with gains of 1.81--3.09 points (4.8--8.3\% relative). The improvements span multiple reasoning benchmarks, although their magnitude varies across tasks and saturation ratios. For example, at 80\% saturation, Negative Rollout substantially improves AIME 2025 (15.21$\rightarrow$30.21) and GPQA (37.88$\rightarrow$41.92), while some individual settings exhibit smaller gains or slight regressions. Importantly, the aggregate benefit persists even when only 40\% of the training data is saturated, where average performance improves from 37.97 to 39.78. These results suggest that Negative Rollout does not require the entire training set to be reward-saturated. Instead, it can be selectively integrated into standard GRPO training: non-saturated groups retain their original relative learning signals, while Negative Rollout restores learning signals only for saturated groups. This makes the intervention applicable to mixed training settings in which saturation occurs only for a subset of prompts.

\subsection{Towards RSI-Style Self-Evolving Saturation-Aware Training}

\begin{wrapfigure}{r}{0.44\linewidth}
    \centering
    \vspace{-15pt}
    \includegraphics[width=1\linewidth]{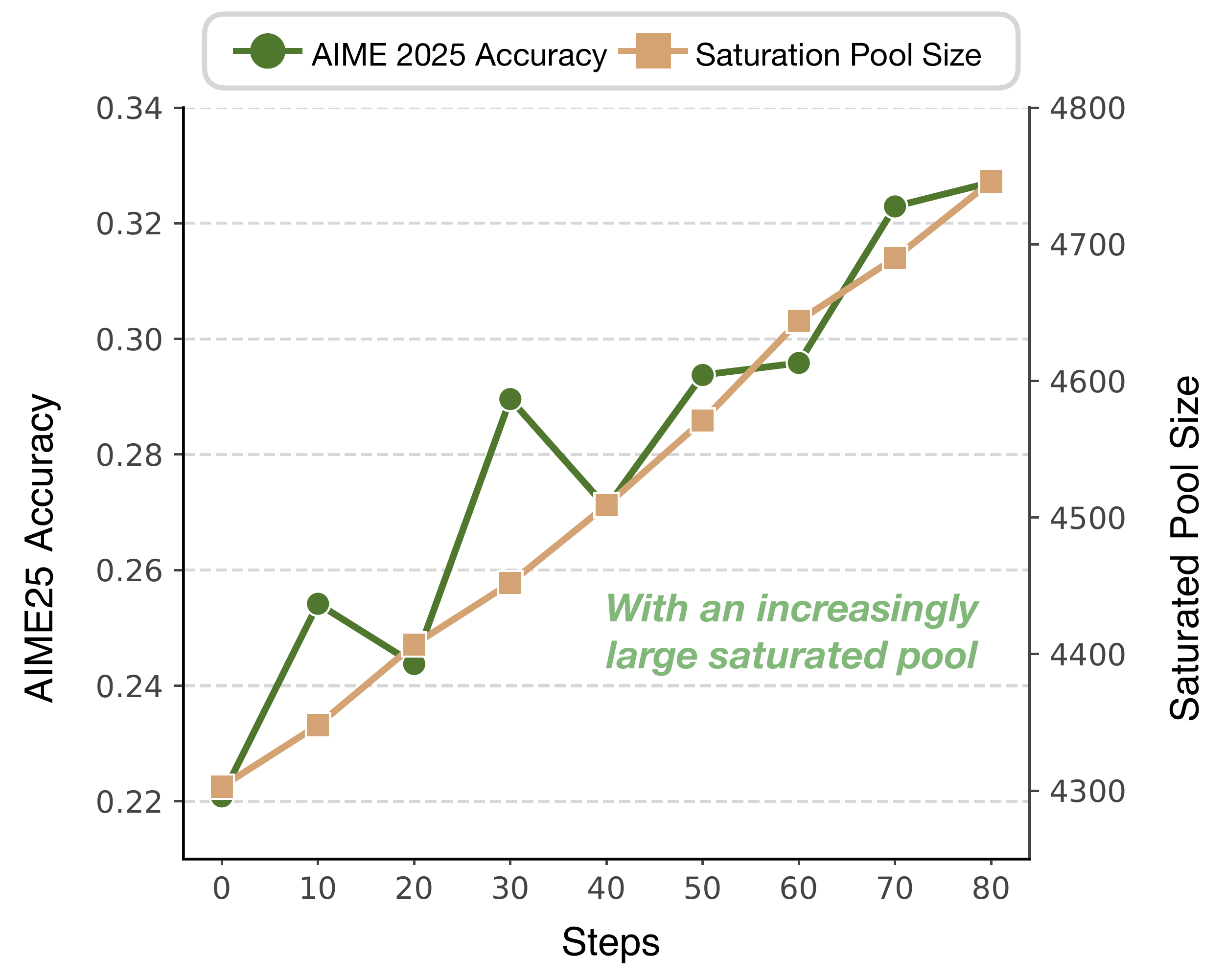}
    \vspace{-15pt}
    \caption{Self-evolving saturation-aware training, with expanding saturated pool and improving AIME25 performance.}    \label{fig:iteration_dynamics}
    \vspace{-15pt}
\end{wrapfigure}

Reward saturation is not necessarily a static property of a dataset: as the policy improves during RL training, previously challenging prompts may become saturated. To capture this dynamic, we periodically identify newly saturated prompts from the remaining data and add them to the saturation pool. As shown in Figure~\ref{fig:iteration_dynamics}, the saturated pool expands by 443 newly discovered prompts by step 80, showing that new reward-saturated prompts continue to emerge as training progresses. 

This iterative process is accompanied by continued performance gains, with validation accuracy on AIME 2025 increasing from 0.2208 at step 0 to 0.3271 at step 80. Rather than discarding these solved examples, negative rollout construction converts them back into prompts with informative learning signals. We refer to this iterative procedure as an RSI-style self-evolving saturation-aware training loop in which policy improvement creates newly saturated prompts, which are identified and converted into informative training signals to further improve the policy. Although our experiment is a preliminary realization of this paradigm, it suggests that training data can evolve together with model capability, reducing the need to continually replace solved examples with newly constructed harder problems.


\subsection{A Larger Advantage doesn't Lead to Better Performance} 

Zero-reward padding provides a direct way to restore non-zero group-relative advantages in fully saturated groups. Consider a saturated group of $G$ successful rollouts with reward $1$. After appending $k$ artificial samples with reward $0$, the advantages for the original correct $A_{+}(k)$ and appended wrong $A_{-}(k)$ rollouts become:

\begin{equation*}
A_{+}(k)
=
\sqrt{
\frac{k(G+k-1)}{G(G+k)}}, \ \
A_{-}(k)
=
-\sqrt{\frac{G(G+k-1)}{k(G+k)}}
\end{equation*}

Notably, $A_{+}(k)$ increases monotonically with $k$ (full derivation in  Appendix~\ref{appendix:add_zero_derivation}). However, all $G$ successful rollouts receive the same $A_{+}(k)$ regardless of their reasoning quality or trajectory structure. Increasing $k$ therefore amplifies the advantage magnitude without providing additional distinctions among successful trajectories.

We empirically examine this effect on Qwen3-4B with $k\in\{1, 2, 3, 8, \text{rand}\}$. As shown in Table~\ref{tab:add_zero}, performance does not improve monotonically with $k$, despite the monotonic increase in $A_{+}(k)$, and all zero-padding variants underperform standard GRPO. This reveals an important distinction between \emph{advantage magnitude} and \emph{advantage informativeness}: zero-reward padding restores non-zero advantages, but the resulting contrast is purely numerical and provides no additional trajectory-level information. Effective recovery from reward saturation therefore requires not only a non-zero learning signal, but also one that meaningfully distinguishes sampled trajectories.

\begin{table*}[h]
\centering
\caption{Effect of number of zero-reward padding size $k$ on Qwen3-4B.}
\vspace{-5pt}
\label{tab:add_zero}
\resizebox{\textwidth}{!}{
\begin{tabular}{lccccccccc}
\toprule[1.75pt]
\textbf{Settings} & \textbf{MATH500} & \textbf{GPQA}  & \textbf{BBH} & \textbf{Minerva} &  \textbf{AIME24} & \textbf{AIME25} & \textbf{IFBench} & \textbf{IFEval} & \textbf{Average} \\
\midrule
GRPO & 70.60 & 31.31 & 16.13 & 25.74 & 31.98 & 27.40 & 28.00 & 78.80 & 38.74 \\
\midrule
Zero Padding ($k=1$) & 68.20 & 34.85 & 15.11 & 23.90 & 21.67 & 19.27 & 29.20 & 80.60 & 36.60 \\
Zero Padding ($k=2$) & 62.80 & 31.82 & 13.83 & 19.85 & 14.06 & 11.88 & 27.60 & 81.40 & 32.90 \\
Zero Padding ($k=3$) & 67.20 & 30.30 & 13.50 & 22.79 & 20.21 & 23.13 & 30.00 & 79.80 & 35.87 \\
Zero Padding ($k=8$) & 62.20 & 29.80 & 13.50 & 22.79 & 17.71 & 3.65 & 29.60 & 79.60 & 32.36 \\
Zero Padding (random $k$) & 66.40 & 30.30 & 13.27 & 22.43 & 17.08 & 18.96 & 27.60 & 80.00 & 34.51 \\
\bottomrule[1.75pt]
\end{tabular}}
\vspace{-5pt}
\end{table*}

\subsection{Generalization Across Model Families}

To examine whether negative rollout construction generalizes beyond Qwen3, we further evaluate it on LLaMA-3.1-8B-Instruct~\citep{grattafiori2024llama3herdmodels}. As shown in Table~\ref{tab:math_benchmark_llama}, GRPO improves the average score from 22.98 to 24.97, while negative rollout further improves it to 25.63, with pronounced gains on Minerva (13.97$\rightarrow$20.59), GPQA (20.71$\rightarrow$24.75), and AIME24 (3.30$\rightarrow$10.00). In contrast, Mixed-CUTS transfers poorly to LLaMA-3.1-8B-Instruct, reducing the average score to 17.35. These results suggest that negative rollouts provide a more robust intervention for reward saturation across model families without model-specific adaptation.

\begin{table*}[h]
\centering
\caption{Generalization results on LLaMA-3.1-8B-Instruct across eight benchmarks.}
\vspace{-5pt}
\label{tab:math_benchmark_llama}
\resizebox{\textwidth}{!}{
\begin{tabular}{lccccccccc}
\toprule[1.75pt]
\textbf{Settings} & \textbf{MATH500} & \textbf{GPQA}  & \textbf{BBH} & \textbf{Minerva} &  \textbf{AIME24} & \textbf{AIME25} & \textbf{IFBench} & \textbf{IFEval} & \textbf{Average} \\
\midrule
Base Model & 34.00 & 18.18 & 8.96 & 12.50 & 6.25 & 1.15 & 30.80 & 72.00 & 22.98 \\
GRPO & 45.00 & 20.71 & 10.42 & 13.97 &3.30 & 1.04	& 32.90 & 72.40 & 24.97 \\
Mixed-CUTS & 27.00 & 3.03 & 2.50 & 9.93 & 3.30 & 0.21 & 29.20 & 63.60 & 17.35 \\
\midrule
Negative Rollout & 43.60 & 24.75 & 9.27 & 20.59 & 10.00 & 0.42& 29.20 & 67.20 & \textbf{25.63} \\
\bottomrule[1.75pt]
\end{tabular}}
\vspace{-5pt}
\end{table*}

\section{Conclusion}

We study reward saturation in GRPO, where all rollouts for a prompt receive the same high reward and therefore provide little relative advantage signal. Across interventions at the data, rollout, reward, and advantage levels, our experiments show that constructing wrong-answer rollouts at the rollout level is the most effective way to recover useful learning signals from saturated examples. The method consistently improves average performance across Qwen3-1.7B and Qwen3-4B, and transfers to LLaMA3.1-8B-Instruct. Our analyses further show that simply changing reward statistics, adding generic perturbations, or imitating teacher solutions is less reliable. Overall, the results suggest that saturated data can remain useful for RL when it is paired with semantically meaningful negative trajectories that reveal plausible reasoning failures.

\subsection*{AI use statement}

In this work, we used generative AI tools for aiding and polishing writing. We did not use generative AI tools to design or provide feedback on research methodology, conduct experiments, or generate synthetic datasets, and the remaining disclosure categories are not applicable to this work. We take responsibility for the final content of this work, including text, claims or artifacts produced with the aid of generative AI. 

\subsection*{Ethics Statement}

This work studies reinforcement learning with reward-saturated data for large language models. Our experiments use public mathematical and general reasoning benchmarks and do not involve human subjects or private data. The negative trajectories introduced by our method are model-generated and used solely as training signals. 

\subsection*{Reproducibility Statement}

We provide extensive experiment details such as hyperparameter settings, dataset statistics, and more in Appendix~\ref{appendix:Experimental_details}. The training and inference code is in \href{https://github.com/Ziyuan-Yang/saturatedRL}{https://github.com/Ziyuan-Yang/saturatedRL}.

\bibliography{iclr2027_conference}

@misc{shao2024deepseekmathpushinglimitsmathematical,
      title={DeepSeekMath: Pushing the Limits of Mathematical Reasoning in Open Language Models}, 
      author={Zhihong Shao and Peiyi Wang and Qihao Zhu and Runxin Xu and Junxiao Song and Xiao Bi and Haowei Zhang and Mingchuan Zhang and Y. K. Li and Y. Wu and Daya Guo},
      year={2024},
      eprint={2402.03300},
      archivePrefix={arXiv},
      primaryClass={cs.CL},
      url={https://arxiv.org/abs/2402.03300}, 
}

@inproceedings{
yu2025dapo,
title={{DAPO}: An Open-Source {LLM} Reinforcement Learning System at Scale},
author={Qiying Yu and Zheng Zhang and Ruofei Zhu and Yufeng Yuan and Xiaochen Zuo and YuYue and Weinan Dai and Tiantian Fan and Gaohong Liu and Juncai Liu and LingJun Liu and Xin Liu and Haibin Lin and Zhiqi Lin and Bole Ma and Guangming Sheng and Yuxuan Tong and Chi Zhang and Mofan Zhang and Ru Zhang and Wang Zhang and Hang Zhu and Jinhua Zhu and Jiaze Chen and Jiangjie Chen and Chengyi Wang and Hongli Yu and Yuxuan Song and Xiangpeng Wei and Hao Zhou and Jingjing Liu and Wei-Ying Ma and Ya-Qin Zhang and Lin Yan and Yonghui Wu and Mingxuan Wang},
booktitle={The Thirty-ninth Annual Conference on Neural Information Processing Systems},
year={2025},
url={https://openreview.net/forum?id=2a36EMSSTp}
}

@inproceedings{
huang2026rzero,
title={R-Zero: Self-Evolving Reasoning {LLM} from Zero Data},
author={Chengsong Huang and Wenhao Yu and Xiaoyang Wang and Hongming Zhang and Zongxia Li and Ruosen Li and Jiaxin Huang and Haitao Mi and Dong Yu},
booktitle={The Fourteenth International Conference on Learning Representations},
year={2026},
url={https://openreview.net/forum?id=96apU6YzSO}
}

@article{deepseekr1,
author = {{DeepSeek-AI}},
title = {DeepSeek-R1 incentivizes reasoning in LLMs through reinforcement learning},
journal = {Nature},
volume  = {645},
pages   = {633--638},
year    = {2025},
doi     = {10.1038/s41586-025-09422-z}
}

@inproceedings{liang2026correctlearnreinforcementlearning,
    title = "Too Correct to Learn: Reinforcement Learning on Saturated Reasoning Data",
    author = "Liang, Zhenwen  and
      Zhou, Yujun  and
      Lu, Sidi  and
      Zhang, Xiangliang  and
      Mi, Haitao  and
      Yu, Dong",
    editor = "Liakata, Maria  and
      Moreira, Viviane P.  and
      Zhang, Jiajun  and
      Jurgens, David",
    booktitle = "Proceedings of the 64th Annual Meeting of the {A}ssociation for {C}omputational {L}inguistics (Volume 2: Short Papers)",
    month = jul,
    year = "2026",
    address = "San Diego, California, United States",
    publisher = "Association for Computational Linguistics",
    url = "https://aclanthology.org/2026.acl-short.19/",
    doi = "10.18653/v1/2026.acl-short.19",
    pages = "205--215",
}

@misc{yang2025qwen3technicalreport,
      title={Qwen3 Technical Report}, 
      author={An Yang and Anfeng Li and Baosong Yang and Beichen Zhang and Binyuan Hui and Bo Zheng and Bowen Yu and Chang Gao and Chengen Huang and Chenxu Lv and Chujie Zheng and Dayiheng Liu and Fan Zhou and Fei Huang and Feng Hu and Hao Ge and Haoran Wei and Huan Lin and Jialong Tang and Jian Yang and Jianhong Tu and Jianwei Zhang and Jianxin Yang and Jiaxi Yang and Jing Zhou and Jingren Zhou and Junyang Lin and Kai Dang and Keqin Bao and Kexin Yang and Le Yu and Lianghao Deng and Mei Li and Mingfeng Xue and Mingze Li and Pei Zhang and Peng Wang and Qin Zhu and Rui Men and Ruize Gao and Shixuan Liu and Shuang Luo and Tianhao Li and Tianyi Tang and Wenbiao Yin and Xingzhang Ren and Xinyu Wang and Xinyu Zhang and Xuancheng Ren and Yang Fan and Yang Su and Yichang Zhang and Yinger Zhang and Yu Wan and Yuqiong Liu and Zekun Wang and Zeyu Cui and Zhenru Zhang and Zhipeng Zhou and Zihan Qiu},
      year={2025},
      eprint={2505.09388},
      archivePrefix={arXiv},
      primaryClass={cs.CL},
      url={https://arxiv.org/abs/2505.09388}, 
}

@inproceedings{
zeng2025simplerl,
title={Simple{RL}-Zoo: Investigating and Taming Zero Reinforcement Learning for Open Base Models in the Wild},
author={Weihao Zeng and Yuzhen Huang and Qian Liu and Wei Liu and Keqing He and Zejun MA and Junxian He},
booktitle={Second Conference on Language Modeling},
year={2025},
url={https://openreview.net/forum?id=vSMCBUgrQj}
}

@inproceedings{
hendrycks2021measuring,
title={Measuring Mathematical Problem Solving With the {MATH} Dataset},
author={Dan Hendrycks and Collin Burns and Saurav Kadavath and Akul Arora and Steven Basart and Eric Tang and Dawn Song and Jacob Steinhardt},
booktitle={Thirty-fifth Conference on Neural Information Processing Systems Datasets and Benchmarks Track (Round 2)},
year={2021},
url={https://openreview.net/forum?id=7Bywt2mQsCe}
}

@inproceedings{rein2024gpqa,
  title={{Gpqa}: A graduate-level google-proof Q\&A benchmark},
  author={Rein, David and Hou, Betty Li and Stickland, Asa Cooper and Petty, Jackson and Pang, Richard Yuanzhe and Dirani, Julien and Michael, Julian and Bowman, Samuel R},
  booktitle={First Conference on Language Modeling},
  year={2024},
  url={https://openreview.net/forum?id=Ti67584b98}

}

@inproceedings{
lewkowycz2022solving,
title={Solving Quantitative Reasoning Problems with Language Models},
author={Aitor Lewkowycz and Anders Johan Andreassen and David Dohan and Ethan Dyer and Henryk Michalewski and Vinay Venkatesh Ramasesh and Ambrose Slone and Cem Anil and Imanol Schlag and Theo Gutman-Solo and Yuhuai Wu and Behnam Neyshabur and Guy Gur-Ari and Vedant Misra},
booktitle={The Thirty-Sixth Annual Conference on Neural Information Processing Systems},
year={2022},
url={https://openreview.net/forum?id=IFXTZERXdM7}
}

@inproceedings{pyatkin2025generalizing,
title={Generalizing Verifiable Instruction Following},
author={Valentina Pyatkin and Saumya Malik and Victoria Graf and Hamish Ivison and Shengyi Huang and Pradeep Dasigi and Nathan Lambert and Hannaneh Hajishirzi},
booktitle={The Thirty-ninth Conference on Neural Information Processing Systems},
year={2025},
url={https://openreview.net/forum?id=yfYgwjj5F8}
}

@misc{zhou2023instructionfollowingevaluationlargelanguage,
title={Instruction-Following Evaluation for Large Language Models}, 
author={Jeffrey Zhou and Tianjian Lu and Swaroop Mishra and Siddhartha Brahma and Sujoy Basu and Yi Luan and Denny Zhou and Le Hou},
year={2023},
eprint={2311.07911},
archivePrefix={arXiv},
primaryClass={cs.CL},
url={https://arxiv.org/abs/2311.07911}, 
}

@inproceedings{suzgun-etal-2023-challenging,
    title = "Challenging {BIG}-Bench Tasks and Whether Chain-of-Thought Can Solve Them",
    author = {Suzgun, Mirac  and
      Scales, Nathan  and
      Sch{\"a}rli, Nathanael  and
      Gehrmann, Sebastian  and
      Tay, Yi  and
      Chung, Hyung Won  and
      Chowdhery, Aakanksha  and
      Le, Quoc  and
      Chi, Ed  and
      Zhou, Denny  and
      Wei, Jason},
    editor = "Rogers, Anna  and
      Boyd-Graber, Jordan  and
      Okazaki, Naoaki",
    booktitle = "Findings of the Association for Computational Linguistics: ACL 2023",
    month = jul,
    year = "2023",
    address = "Toronto, Canada",
    publisher = "Association for Computational Linguistics",
    url = "https://aclanthology.org/2023.findings-acl.824/",
    doi = "10.18653/v1/2023.findings-acl.824",
    pages = "13003--13051",
}

@inproceedings{sheng2024hybridflow,
author = {Sheng, Guangming and Zhang, Chi and Ye, Zilingfeng and Wu, Xibin and Zhang, Wang and Zhang, Ru and Peng, Yanghua and Lin, Haibin and Wu, Chuan},
title = {HybridFlow: A Flexible and Efficient RLHF Framework},
year = {2025},
url = {https://doi.org/10.1145/3689031.3696075},
doi = {10.1145/3689031.3696075},
booktitle = {Proceedings of the Twentieth European Conference on Computer Systems},
pages = {1279–1297},
}

@dataset{aime_1983_2024,
  author = {Hemish Veeraboina},
  title = {AIME Problem Set 1983-2024},
  year = {2023},
  publisher = {Kaggle},
  url = {https://www.kaggle.com/datasets/hemishveeraboina/aime-problem-set-1983-2024}
}

@misc{zhang2026aero,
title={Train Less, Learn More: Adaptive Efficient Rollout Optimization for Group-Based Reinforcement Learning}, 
author={Zhi Zhang and Zhen Han and Costas Mavromatis and Qi Zhu and Yunyi Zhang and Sheng Guan and Dingmin Wang and Xiong Zhou and Shuai Wang and Soji Adeshina and Vassilis Ioannidis and Huzefa Rangwala},
year={2026},
eprint={2602.14338},
archivePrefix={arXiv},
primaryClass={cs.LG},
url={https://arxiv.org/abs/2602.14338}, 
}

@inproceedings{
huang26multiVIGOR,
title={Learning as Reasoning Unfolds: Progressive Rollout Allocation for Efficient Reinforcement Learning},
author={Heyang Jiang and Henry Liu and Baharan Mirzasoleiman},
booktitle={Third Conference on Language Modeling},
year={2026},
url={https://openreview.net/forum?id=qoJJK90DGh}
}

@misc{agrawal2026offcontext,
title={Off-Context GRPO: Learning to Reason on Hard Problems using Privileged Information}, 
author={Priyank Agrawal and Ankur Samanta and Shervin Ghasemlou and Jalaj Bhandari and Kavosh Asadi and Daniel Jiang and Aditya Modi},
year={2026},
eprint={2607.19313},
archivePrefix={arXiv},
primaryClass={cs.LG},
url={https://arxiv.org/abs/2607.19313}, 
}

@inproceedings{
he2026advantage,
title={Advantage Collapse in Group Relative Policy Optimization: Diagnosis and Mitigation},
author={Xixiang He and Qiyao Sun and Ao Cheng and Xingming Li and Xuanyu Ji and Hailun Lu and Runke Huang and Qingyong Hu},
booktitle={Forty-third International Conference on Machine Learning},
year={2026},
url={https://openreview.net/forum?id=MKNimf9bIx}
}

@inproceedings{
le2026no,
title={No Prompt Left Behind: Exploiting Zero-Variance Prompts in {LLM} Reinforcement Learning via Entropy-Guided Advantage Shaping},
author={Thanh-Long V. Le and Myeongho Jeon and Kim Vu and Viet Dac Lai and Eunho Yang},
booktitle={The Fourteenth International Conference on Learning Representations},
year={2026},
url={https://openreview.net/forum?id=kiXFIESZKv}
}

@inproceedings{
zhao2025absolute,
title={Absolute Zero: Reinforced Self-play Reasoning with Zero Data},
author={Andrew Zhao and Yiran Wu and Yang Yue and Tong Wu and Quentin Xu and Yang Yue and Matthieu Lin and Shenzhi Wang and Qingyun Wu and Zilong Zheng and Gao Huang},
booktitle={The Thirty-ninth Annual Conference on Neural Information Processing Systems},
year={2025},
url={https://openreview.net/forum?id=neZSGqhxDa}
}

@inproceedings{
parashar2026curriculum,
title={Curriculum Reinforcement Learning from Easy to Hard Tasks Improves {LLM} Reasoning},
author={Shubham Parashar and Shurui Gui and Xiner Li and Hongyi Ling and Sushil Vemuri and Blake Olson and Eric Li and Yu Zhang and James Caverlee and Dileep Kalathil and Shuiwang Ji},
booktitle={The Fourteenth International Conference on Learning Representations},
year={2026},
url={https://openreview.net/forum?id=KJvHnl3kUv}
}

@misc{bao2026questionbeget,
title={Question Begets Question: Self-Evolving Curriculum for Reinforcement Fine-Tuning on Competition Mathematics}, 
author={Longtian Bao and Jianyou Wang and Yang Zhang and Youze Zheng and Ramamohan Paturi},
year={2026},
eprint={2608.01522},
archivePrefix={arXiv},
primaryClass={cs.LG},
url={https://arxiv.org/abs/2608.01522}, 
}

@misc{grattafiori2024llama3herdmodels,
      title={The Llama 3 Herd of Models}, 
      author={Aaron Grattafiori and Abhimanyu Dubey and Abhinav Jauhri and Abhinav Pandey and Abhishek Kadian and Ahmad Al-Dahle and Aiesha Letman and Akhil Mathur and Alan Schelten and Alex Vaughan and Amy Yang and Angela Fan and Anirudh Goyal and Anthony Hartshorn and Aobo Yang and Archi Mitra and Archie Sravankumar and Artem Korenev and Arthur Hinsvark and Arun Rao and Aston Zhang and Aurelien Rodriguez and Austen Gregerson and Ava Spataru and Baptiste Roziere and Bethany Biron and Binh Tang and Bobbie Chern and Charlotte Caucheteux and Chaya Nayak and Chloe Bi and Chris Marra and Chris McConnell and Christian Keller and Christophe Touret and Chunyang Wu and Corinne Wong and Cristian Canton Ferrer and Cyrus Nikolaidis and Damien Allonsius and Daniel Song and Danielle Pintz and Danny Livshits and Danny Wyatt and David Esiobu and Dhruv Choudhary and Dhruv Mahajan and Diego Garcia-Olano and Diego Perino and Dieuwke Hupkes and Egor Lakomkin and Ehab AlBadawy and Elina Lobanova and Emily Dinan and Eric Michael Smith and Filip Radenovic and Francisco Guzmán and Frank Zhang and Gabriel Synnaeve and Gabrielle Lee and Georgia Lewis Anderson and Govind Thattai and Graeme Nail and Gregoire Mialon and Guan Pang and Guillem Cucurell and Hailey Nguyen and Hannah Korevaar and Hu Xu and Hugo Touvron and Iliyan Zarov and Imanol Arrieta Ibarra and Isabel Kloumann and Ishan Misra and Ivan Evtimov and Jack Zhang and Jade Copet and Jaewon Lee and Jan Geffert and Jana Vranes and Jason Park and Jay Mahadeokar and Jeet Shah and Jelmer van der Linde and Jennifer Billock and Jenny Hong and Jenya Lee and Jeremy Fu and Jianfeng Chi and Jianyu Huang and Jiawen Liu and Jie Wang and Jiecao Yu and Joanna Bitton and Joe Spisak and Jongsoo Park and Joseph Rocca and Joshua Johnstun and Joshua Saxe and Junteng Jia and Kalyan Vasuden Alwala and Karthik Prasad and Kartikeya Upasani and Kate Plawiak and Ke Li and Kenneth Heafield and Kevin Stone and Khalid El-Arini and Krithika Iyer and Kshitiz Malik and Kuenley Chiu and Kunal Bhalla and Kushal Lakhotia and Lauren Rantala-Yeary and Laurens van der Maaten and Lawrence Chen and Liang Tan and Liz Jenkins and Louis Martin and Lovish Madaan and Lubo Malo and Lukas Blecher and Lukas Landzaat and Luke de Oliveira and Madeline Muzzi and Mahesh Pasupuleti and Mannat Singh and Manohar Paluri and Marcin Kardas and Maria Tsimpoukelli and Mathew Oldham and Mathieu Rita and Maya Pavlova and Melanie Kambadur and Mike Lewis and Min Si and Mitesh Kumar Singh and Mona Hassan and Naman Goyal and Narjes Torabi and Nikolay Bashlykov and Nikolay Bogoychev and Niladri Chatterji and Ning Zhang and Olivier Duchenne and Onur Çelebi and Patrick Alrassy and Pengchuan Zhang and Pengwei Li and Petar Vasic and Peter Weng and Prajjwal Bhargava and Pratik Dubal and Praveen Krishnan and Punit Singh Koura and Puxin Xu and Qing He and Qingxiao Dong and Ragavan Srinivasan and Raj Ganapathy and Ramon Calderer and Ricardo Silveira Cabral and Robert Stojnic and Roberta Raileanu and Rohan Maheswari and Rohit Girdhar and Rohit Patel and Romain Sauvestre and Ronnie Polidoro and Roshan Sumbaly and Ross Taylor and Ruan Silva and Rui Hou and Rui Wang and Saghar Hosseini and Sahana Chennabasappa and Sanjay Singh and Sean Bell and Seohyun Sonia Kim and Sergey Edunov and Shaoliang Nie and Sharan Narang and Sharath Raparthy and Sheng Shen and Shengye Wan and Shruti Bhosale and Shun Zhang and Simon Vandenhende and Soumya Batra and Spencer Whitman and Sten Sootla and Stephane Collot and Suchin Gururangan and Sydney Borodinsky and Tamar Herman and Tara Fowler and Tarek Sheasha and Thomas Georgiou and Thomas Scialom and Tobias Speckbacher and Todor Mihaylov and Tong Xiao and Ujjwal Karn and Vedanuj Goswami and Vibhor Gupta and Vignesh Ramanathan and Viktor Kerkez and Vincent Gonguet and Virginie Do and Vish Vogeti and Vítor Albiero and Vladan Petrovic and Weiwei Chu and Wenhan Xiong and Wenyin Fu and Whitney Meers and Xavier Martinet and Xiaodong Wang and Xiaofang Wang and Xiaoqing Ellen Tan and Xide Xia and Xinfeng Xie and Xuchao Jia and Xuewei Wang and Yaelle Goldschlag and Yashesh Gaur and Yasmine Babaei and Yi Wen and Yiwen Song and Yuchen Zhang and Yue Li and Yuning Mao and Zacharie Delpierre Coudert and Zheng Yan and Zhengxing Chen and Zoe Papakipos and Aaditya Singh and Aayushi Srivastava and Abha Jain and Adam Kelsey and Adam Shajnfeld and Adithya Gangidi and Adolfo Victoria and Ahuva Goldstand and Ajay Menon and Ajay Sharma and Alex Boesenberg and Alexei Baevski and Allie Feinstein and Amanda Kallet and Amit Sangani and Amos Teo and Anam Yunus and Andrei Lupu and Andres Alvarado and Andrew Caples and Andrew Gu and Andrew Ho and Andrew Poulton and Andrew Ryan and Ankit Ramchandani and Annie Dong and Annie Franco and Anuj Goyal and Aparajita Saraf and Arkabandhu Chowdhury and Ashley Gabriel and Ashwin Bharambe and Assaf Eisenman and Azadeh Yazdan and Beau James and Ben Maurer and Benjamin Leonhardi and Bernie Huang and Beth Loyd and Beto De Paola and Bhargavi Paranjape and Bing Liu and Bo Wu and Boyu Ni and Braden Hancock and Bram Wasti and Brandon Spence and Brani Stojkovic and Brian Gamido and Britt Montalvo and Carl Parker and Carly Burton and Catalina Mejia and Ce Liu and Changhan Wang and Changkyu Kim and Chao Zhou and Chester Hu and Ching-Hsiang Chu and Chris Cai and Chris Tindal and Christoph Feichtenhofer and Cynthia Gao and Damon Civin and Dana Beaty and Daniel Kreymer and Daniel Li and David Adkins and David Xu and Davide Testuggine and Delia David and Devi Parikh and Diana Liskovich and Didem Foss and Dingkang Wang and Duc Le and Dustin Holland and Edward Dowling and Eissa Jamil and Elaine Montgomery and Eleonora Presani and Emily Hahn and Emily Wood and Eric-Tuan Le and Erik Brinkman and Esteban Arcaute and Evan Dunbar and Evan Smothers and Fei Sun and Felix Kreuk and Feng Tian and Filippos Kokkinos and Firat Ozgenel and Francesco Caggioni and Frank Kanayet and Frank Seide and Gabriela Medina Florez and Gabriella Schwarz and Gada Badeer and Georgia Swee and Gil Halpern and Grant Herman and Grigory Sizov and Guangyi and Zhang and Guna Lakshminarayanan and Hakan Inan and Hamid Shojanazeri and Han Zou and Hannah Wang and Hanwen Zha and Haroun Habeeb and Harrison Rudolph and Helen Suk and Henry Aspegren and Hunter Goldman and Hongyuan Zhan and Ibrahim Damlaj and Igor Molybog and Igor Tufanov and Ilias Leontiadis and Irina-Elena Veliche and Itai Gat and Jake Weissman and James Geboski and James Kohli and Janice Lam and Japhet Asher and Jean-Baptiste Gaya and Jeff Marcus and Jeff Tang and Jennifer Chan and Jenny Zhen and Jeremy Reizenstein and Jeremy Teboul and Jessica Zhong and Jian Jin and Jingyi Yang and Joe Cummings and Jon Carvill and Jon Shepard and Jonathan McPhie and Jonathan Torres and Josh Ginsburg and Junjie Wang and Kai Wu and Kam Hou U and Karan Saxena and Kartikay Khandelwal and Katayoun Zand and Kathy Matosich and Kaushik Veeraraghavan and Kelly Michelena and Keqian Li and Kiran Jagadeesh and Kun Huang and Kunal Chawla and Kyle Huang and Lailin Chen and Lakshya Garg and Lavender A and Leandro Silva and Lee Bell and Lei Zhang and Liangpeng Guo and Licheng Yu and Liron Moshkovich and Luca Wehrstedt and Madian Khabsa and Manav Avalani and Manish Bhatt and Martynas Mankus and Matan Hasson and Matthew Lennie and Matthias Reso and Maxim Groshev and Maxim Naumov and Maya Lathi and Meghan Keneally and Miao Liu and Michael L. Seltzer and Michal Valko and Michelle Restrepo and Mihir Patel and Mik Vyatskov and Mikayel Samvelyan and Mike Clark and Mike Macey and Mike Wang and Miquel Jubert Hermoso and Mo Metanat and Mohammad Rastegari and Munish Bansal and Nandhini Santhanam and Natascha Parks and Natasha White and Navyata Bawa and Nayan Singhal and Nick Egebo and Nicolas Usunier and Nikhil Mehta and Nikolay Pavlovich Laptev and Ning Dong and Norman Cheng and Oleg Chernoguz and Olivia Hart and Omkar Salpekar and Ozlem Kalinli and Parkin Kent and Parth Parekh and Paul Saab and Pavan Balaji and Pedro Rittner and Philip Bontrager and Pierre Roux and Piotr Dollar and Polina Zvyagina and Prashant Ratanchandani and Pritish Yuvraj and Qian Liang and Rachad Alao and Rachel Rodriguez and Rafi Ayub and Raghotham Murthy and Raghu Nayani and Rahul Mitra and Rangaprabhu Parthasarathy and Raymond Li and Rebekkah Hogan and Robin Battey and Rocky Wang and Russ Howes and Ruty Rinott and Sachin Mehta and Sachin Siby and Sai Jayesh Bondu and Samyak Datta and Sara Chugh and Sara Hunt and Sargun Dhillon and Sasha Sidorov and Satadru Pan and Saurabh Mahajan and Saurabh Verma and Seiji Yamamoto and Sharadh Ramaswamy and Shaun Lindsay and Shaun Lindsay and Sheng Feng and Shenghao Lin and Shengxin Cindy Zha and Shishir Patil and Shiva Shankar and Shuqiang Zhang and Shuqiang Zhang and Sinong Wang and Sneha Agarwal and Soji Sajuyigbe and Soumith Chintala and Stephanie Max and Stephen Chen and Steve Kehoe and Steve Satterfield and Sudarshan Govindaprasad and Sumit Gupta and Summer Deng and Sungmin Cho and Sunny Virk and Suraj Subramanian and Sy Choudhury and Sydney Goldman and Tal Remez and Tamar Glaser and Tamara Best and Thilo Koehler and Thomas Robinson and Tianhe Li and Tianjun Zhang and Tim Matthews and Timothy Chou and Tzook Shaked and Varun Vontimitta and Victoria Ajayi and Victoria Montanez and Vijai Mohan and Vinay Satish Kumar and Vishal Mangla and Vlad Ionescu and Vlad Poenaru and Vlad Tiberiu Mihailescu and Vladimir Ivanov and Wei Li and Wenchen Wang and Wenwen Jiang and Wes Bouaziz and Will Constable and Xiaocheng Tang and Xiaojian Wu and Xiaolan Wang and Xilun Wu and Xinbo Gao and Yaniv Kleinman and Yanjun Chen and Ye Hu and Ye Jia and Ye Qi and Yenda Li and Yilin Zhang and Ying Zhang and Yossi Adi and Youngjin Nam and Yu and Wang and Yu Zhao and Yuchen Hao and Yundi Qian and Yunlu Li and Yuzi He and Zach Rait and Zachary DeVito and Zef Rosnbrick and Zhaoduo Wen and Zhenyu Yang and Zhiwei Zhao and Zhiyu Ma},
      year={2024},
      eprint={2407.21783},
      archivePrefix={arXiv},
      primaryClass={cs.AI},
      url={https://arxiv.org/abs/2407.21783}, 
}

@misc{mao2026popo,
      title={RLVR without Ineffective Samples: Group Prioritized Off-Policy Optimization for LLM Reasoning}, 
      author={Yixiu Mao and Yun Qu and Qi Wang and Heming Zou and Xiangyang Ji},
      year={2026},
      eprint={2606.01281},
      archivePrefix={arXiv},
      primaryClass={cs.LG},
      url={https://arxiv.org/abs/2606.01281}, 
}

@misc{deng2026prismgrpo,
      title={Prism-GRPO: Faster VLA Policy Optimization via Splitting Same-outcome Groups}, 
      author={Zeyun Deng and Yuzhe Lu and Yawei Wang and Linbo Liu and Qing Ping and Han Ding and Guande Wu and Panpan Xu and Jun Huan},
      year={2026},
      eprint={2608.17423},
      archivePrefix={arXiv},
      primaryClass={cs.RO},
      url={https://arxiv.org/abs/2608.17423}, 
}

@inproceedings{hu2026vade,
  title={VADE: Variance-Aware Dynamic Sampling via Online Sample-Level Difficulty Estimation for Multimodal Reinforcement Learning},
  author={Hu, Zengjie and Qiu, Jiantao and Bai, Tianyi and Yang, Haojin and Yuan, Binhang and Jing, Qi and He, Conghui and Zhang, Wentao},
  booktitle = {Proceedings of the IEEE/CVF Conference on Computer Vision and Pattern Recognition (CVPR) Findings},
  month     = {June},
  year={2026},
  pages     = {9846-9855}
}
\bibliographystyle{iclr2027_conference}

\newpage
\appendix

\section{Limitations}

Our study has several limitations. First, the experiments mainly focus on mathematical and reasoning benchmarks with verifiable rewards. It remains unclear whether the same saturation-aware interventions transfer to open-ended tasks where correctness is harder to verify. Second, our systematic experiments use Qwen3 and LLaMA3.1, while broader scaling studies across larger model families are further needed. Third, wrong-answer rollouts depend on the quality and diversity of the generated incorrect trajectories. Poorly constructed negatives may introduce noise or encourage superficial contrast rather than robust reasoning. Finally, our iterative saturation mining results are preliminary and run for a limited number of steps. Future work should study longer training runs, stronger format control, and more adaptive criteria for deciding when newly saturated prompts should be added back into training.

\section{Experimental Details}
\label{appendix:Experimental_details}

\subsection{System Prompt}
For all experiments, we used the following system prompt to guide the model's generation format, ensuring that it produces a step-by-step reasoning process and a clearly marked final answer \citep{zeng2025simplerl}:
\begin{bluebox}[System Prompt]
\texttt{
Please reason step by step, and put your final answer within \textbackslash boxed\{\}.
}
\end{bluebox}

\subsection{Wrong Answer Prompt}
\label{appendix:wrong_answer_suffix}
\begin{bluebox}[Wrong Answer Suffix]
\texttt{
Now solve the problem as a student with one specific, subtle, and plausible mathematical misconception. Privately choose the misconception first, but do not explicitly mention it. Write a coherent step-by-step solution that consistently follows this misconception, keep the final answer in \textbackslash boxed\{\}, and make the final answer incorrect.
}
\end{bluebox}






\subsection{Hyperparameter Settings}
\label{app:hyperparameters}

We utilize GRPO for post-training. The model is optimized using $\mathrm{AdamW}$ with a learning rate of $1 \times 10^{-6}$ and a weight decay of $1 \times 10^{-2}$. The training batch size is set to 128, and the maximum response length is 4096. For GRPO-specific configurations, we use a group size of $G=8$ and enable KL regularization with coefficient $\beta=1 \times 10^{-3}$. Unless otherwise specified, rollouts are sampled with a temperature of 1.0, and models are trained for 2 epochs.

For intervention-specific configurations, the higher-temperature variant uses a sampling temperature of $t=1.5$, while negative rollout construction uses $G_{\mathrm{wrong}}=2$ constructed negative trajectories per saturated group. For reward-level interventions, we use $\mathtt{Skywork/Skywork\text{-}Reward\text{-}V2\text{-}Qwen3\text{-}8B}$ as the reward model and $\mathtt{Qwen/Qwen3\text{-}8B}$ as the LLM judge. For iterative saturation mining, every 10 training steps, we resample $G=8$ rollouts from the current policy on the remaining training prompts and add prompts whose rollouts all receive correct rewards to the saturated pool. Table~\ref{tab:hyperparameters} summarizes the main hyperparameters.

\begin{table*}[h]
\small
\centering
\renewcommand{\arraystretch}{1.1}
\setlength{\tabcolsep}{6pt}
\caption{Key hyperparameters used.}
\vspace{1mm}
\label{tab:hyperparameters}
\begin{tabular}{ll}
\toprule[1.75pt]
\textbf{Category} & \textbf{Value} \\
\midrule
Group Size ($G$) & 8 \\
Training Batch Size & 128 \\
PPO Mini-batch Size & 32 \\
KL Loss Type & Low-variance KL \\
KL Coefficient ($\beta$) & $1\times10^{-3}$ \\
Rollout Top-$p$ & 1.0 \\
Rollout Temperature & 1.0 \\
Validation Top-$p$ & 0.95 \\
Validation Temperature & 1.0 \\
Learning Rate & $1\times10^{-6}$ \\
Weight Decay & $1\times10^{-2}$ \\
Max Response Length & 4096 \\ 
Epochs & 2 \\
$\lambda$ & 0.5 \\
\bottomrule[1.75pt]
\end{tabular}
\end{table*}

\section{Robustness Across Random Seeds}
\label{appendix:seed}


We further evaluate the robustness of Negative Rollout to training randomness. Specifically, we train Qwen3-1.7B using both standard GRPO and Negative Rollout with three random seeds and report the results in Table~\ref{tab:std}. Negative Rollout consistently outperforms GRPO across all three runs, improving the average score from $29.38\pm0.41$ to $32.04\pm0.60$. This consistent improvement across independent runs suggests that the gains from Negative Rollout are robust to training randomness rather than being specific to a single training seed.

\begin{table*}[t]
\caption{Robustness across three random seeds on Qwen3-1.7B. We report individual runs together with the mean and standard deviation across seeds.}
\vspace{1mm}
\label{tab:std}
\resizebox{\textwidth}{!}{
\begin{tabular}{lccccccccc}
\toprule[1.75pt]
\textbf{Settings} & \textbf{MATH500} & \textbf{GPQA} & \textbf{BBH} & \textbf{Minerva}  &  \textbf{AIME24} & \textbf{AIME25} & \textbf{IFBench} & \textbf{IFEval} & \textbf{Avg.} \\
\midrule
\textit{GRPO} \\
Seed 1 & 62.60 & 24.24 & 10.86 & 17.28  & 17.08 & 9.17 & 21.20 & 69.00 & 28.93 \\
Seed 2 & 64.00 & 22.22 & 12.72 & 16.18  & 19.90 & 10.73 & 22.00 & 68.00 & 29.47 \\
Seed 3 & 63.60 & 23.23 & 12.43 & 17.28  & 19.38 & 8.02  & 24.40 & 69.60 & 29.74 \\
\textbf{Mean $\pm$ Std.} & 63.40 $\pm$ 0.72 & 23.23 $\pm$ 1.01 & 12.00 $\pm$ 1.00 & 16.91 $\pm$ 0.64 & 18.79 $\pm$ 1.50 & 9.31 $\pm$ 1.36 & 22.53 $\pm$ 1.67 & 68.87 $\pm$ 0.81 & 29.38 $\pm$ 0.41\\
\midrule
\textit{Negative Rollout} \\
Seed 1 & 64.80 & 29.80 & 13.01 & 18.38  & 20.31 & 14.38 & 24.00 & 67.60 & 31.54 \\
Seed 2 & 65.20 & 31.31 & 12.52 & 18.01  & 21.68 & 14.90 & 24.80 & 66.60 & 31.88 \\
Seed 3 & 65.80 & 30.81 & 13.05 & 19.85  & 23.96 & 13.96 & 26.00 & 68.20 & 32.70 \\
\textbf{Mean $\pm$ Std.} & 65.27 $\pm$ 0.50 & 30.64 $\pm$ 0.77 & 12.86 $\pm$ 0.30 & 18.75 $\pm$ 0.97 & 21.98 $\pm$ 1.84 & 14.41 $\pm$ 0.47 & 24.93 $\pm$ 1.01 & 67.47 $\pm$ 0.81 & 32.04 $\pm$  0.60 \\
\bottomrule[1.75pt]
\end{tabular}}
\end{table*}


\section{Further Analysis of Supervised Fine-Tuning}
\label{app:sft_analysis}

As shown in Table~\ref{tab:math_benchmark}, SFT and SFT$\rightarrow$GRPO do not consistently outperform standard GRPO, whereas Negative Rollout achieves stronger aggregate performance. To better understand this difference, we examine how SFT changes individual predictions. On MATH, SFT corrects 25 previously incorrect examples but also breaks 25 previously correct ones. On GPQA, it fixes 51 examples while breaking 51, and on AIME24, it fixes 33 while breaking 67. Thus, although SFT substantially changes model behavior, these changes do not consistently translate into improved reasoning accuracy.

One possible explanation is that SFT provides only positive imitation targets. In saturated settings, the model already produces successful trajectories, so additional teacher trajectories provide limited information about which behaviors should be strengthened or avoided. In contrast, Negative Rollout introduces both successful and unsuccessful trajectories for the same prompt, providing trajectory-level distinctions that are absent from positive-only imitation.

\section{A Larger Advantage doesn't Lead to Better Performance} 
\label{appendix:add_zero_derivation}

\subsection{Full Derivation}

We derive the effect of adding zero-reward samples to a saturated GRPO group. Consider a group where all original rollout rewards are correct:
\begin{equation}
    \underbrace{1,1,\ldots,1}_{G\ \mathrm{samples}}.
\end{equation}
The group has zero variance, so standard group normalization produces no useful relative advantage. We append $k$ artificial zero-reward samples:
\begin{equation}
    \underbrace{1,1,\ldots,1}_{G\ \mathrm{samples}},
    \underbrace{0,0,\ldots,0}_{k\ \mathrm{samples}}.
\end{equation}
Let $n=G+k$. Ignoring the small numerical $\epsilon$ in the denominator, the group mean is
\begin{equation}
    \mu = \frac{G}{G+k}.
\end{equation}
Using the sample standard deviation with denominator $n-1$, the variance is
\begin{align}
    \sigma^2
    &= \frac{1}{n-1}
    \left[
    G(1-\mu)^2 + k(0-\mu)^2
    \right] \\
    &= \frac{1}{n-1}
    \left[
    G\left(\frac{k}{n}\right)^2
    + k\left(\frac{G}{n}\right)^2
    \right] \\
    &= \frac{Gk}{(G+k)(G+k-1)}.
\end{align}
Thus,
\begin{equation}
    \sigma =
    \sqrt{
    \frac{Gk}{(G+k)(G+k-1)}
    }.
\end{equation}

For an original reward-1 sample, the normalized advantage is
\begin{align}
    A_1(k)
    = \frac{1-\mu}{\sigma} 
    &= \frac{\frac{k}{G+k}}
    {
    \sqrt{\frac{Gk}{(G+k)(G+k-1)}}
    } 
    =
    \sqrt{
    \frac{k(G+k-1)}{G(G+k)}
    }.
\end{align}
For an added zero-reward sample, the normalized advantage is
\begin{equation}
    A_0(k)
    =
    -\sqrt{
    \frac{G(G+k-1)}{k(G+k)}
    }.
\end{equation}

The per-correct-sample advantage $A_1(k)$ is monotonically increasing in $k$. To see this, maximize its square and ignore the constant $1/G$:
\begin{equation}
    f(k)=\frac{k(G+k-1)}{G+k}
    = k-\frac{k}{G+k}.
\end{equation}
Then
\begin{equation}
    f'(k)=1-\frac{G}{(G+k)^2} > 0
\end{equation}
for $G>1$ and $k\geq 0$. Therefore, maximizing per-sample advantage alone would suggest adding arbitrarily many zero-reward samples.

\end{document}